 \documentclass[final,5p,times,twocolumn, compress]{elsarticle}

\usepackage{times}
\usepackage{epsfig}
\usepackage{graphicx}
\usepackage{float}

\usepackage{amsmath}
\usepackage {mathtools}
\usepackage[ruled,vlined]{algorithm2e}
\usepackage[breaklinks=true,bookmarks=false]{hyperref}
\usepackage[table, xcdraw]{xcolor}
\definecolor{commentcolor}{RGB}{110,154,155}   
\newcommand{\PyComment}[1]{\ttfamily\textcolor{commentcolor}{\# #1}}  
\newcommand{\PyCode}[1]{\ttfamily\textcolor{black}{#1}} 

\usepackage{amssymb}
\usepackage{lipsum}

\journal{Elsevier}

\begin{document}
\begin{frontmatter}

\title{Multi-network Contrastive Learning Based on Global and Local Features}
\author{Weiquan Li}
\author{Xianzhong Long\corref{mycorrespondingauthor}}
\cortext[mycorrespondingauthor]{Corresponding author}
\ead{lxz@njupt.edu.cn}
\author{Yun Li}
\address{School of Computer Science, Nanjing University of Posts and Telecommunications, Nanjing, 210023, China}




\begin{abstract}
    The popularity of self-supervised learning has made it possible to train models without relying on labeled data, which saves expensive annotation costs. However, most existing self-supervised contrastive learning methods often overlook the combination of global and local feature information. This paper proposes a multi-network contrastive learning framework based on global and local features (MNCLGLF). Global features and local features are the features extracted by the convolutional network from the entire image and specific regions of the image respectively. The model learns feature information at different scales of an image by contrasting the embedding pairs generated by multiple networks. The framework also expands the number of samples used for contrast and improves the training efficiency of the model. Linear evaluation results on three benchmark datasets show that MNCLGLF outperforms several existing classical self-supervised learning methods.
\end{abstract}

\end{frontmatter}

\section{Introduction}

The training of deep neural network models typically requires large amounts of labeled data, which can be costly to obtain. In recent years, self-supervised learning has emerged as a popular research direction in the field of deep learning. One of its key advantages is the ability to leverage large amounts of unlabeled data to train models. Self-supervised learning methods aim to learn useful representations from data without the need for manual annotation. These methods often involve defining a pretext task that can be solved using only the data itself. By training a model to solve this task, it can learn features that are useful for downstream tasks.

Contrastive learning is a common self-supervised learning method that has shown promising results in visual tasks. It uses instance discrimination \cite{aut2, aut1, aut3, aut4} as a pretext task to train models. The usual practice is to use two augmented views of an image as a positive pair. Negative samples come from views obtained by augmenting other images. The pretext task is completed by pulling positive pairs in the embedding space while pushing away negative samples. In the process of training the model to complete the pretext task, the model can learn features that are truly useful for downstream tasks. For example, MoCo \cite{aut2} and SimCLR \cite{aut1} both perform contrastive learning according to the above process to train models and achieve good results in downstream tasks. In addition, BYOL \cite{aut5} can complete contrastive learning without negative samples. Due to its asymmetric structure, SimSiam \cite{aut6} can ensure that the model does not collapse during training. SwAV \cite{aut7} supports the model to complete contrastive learning tasks through clustering.

However, most existing contrastive learning methods only use image-level feature information and ignore the correlation between global and local features. Typically, local feature information is important in dense prediction tasks, since these tasks rely on models operating at the pixel or patch level \cite{aut40}. If only global feature information is used, it is difficult for the model to learn features capable of accomplishing these tasks. For example, the goal of semantic segmentation \cite{aut41} is to associate each pixel in the image with a specific semantic label, while the goal of object detection \cite{aut42} is to determine the location and category of different objects present in the image. Therefore, adding local feature information is more conducive to the model to complete these pixel-level or patch-level prediction tasks. But this does not mean that local feature information is no longer important in downstream tasks such as image-level prediction \cite{aut43}. Since local feature information can enhance the performance of models in dense prediction tasks, it can help the model learn information that cannot be noticed when only using global feature information. Therefore, better methods are needed to comprehensively use global and local feature information.

This paper proposes a multi-network contrastive learning framework. It employs multiple networks to comprehensively use global and local feature information for contrastive learning. Our method can correlate local features with global features, establish relationships between feature information at different scales, and enhance the generalization performance of the model. At the same time, this framework can also provide more samples for contrastive learning, which can improve the performance and training efficiency of the model.
Our contributions are as follows:

\begin{itemize}
\item We introduced global and local feature information into self-supervised contrastive learning, which is achieved through multiple network branches. In one of the network branches, a single view generates multiple patches using a specified dividing strategy. Local feature information can be extracted from these patches. The other network branches extract global feature information from the overall image. Therefore, the model can effectively utilize feature information at different scales to learn better representations.
\item With the same training data, we generated more samples for contrastive learning. Multiple network branches increase the quantity of samples for contrast. Additionally, through the dividing operation, two views can provide multiple positive pairs. The more samples involved in the contrastive learning, the higher the model's performance and training efficiency.
\item We propose a multi-network contrastive learning framework based on global and local features (MNCLGLF), consisting of three network branches. The model's learning process is completed by contrasting the outputs of these three network branches. Our method demonstrates good generalization performance on three benchmark datasets.
\end{itemize}

\section{Related Work}
\subsection{Self-Supervised Learning}
Supervised learning has been the most popular method in computer vision and natural language processing in the past. This method requires a large amount of labeled data, which can be expensive. However, self-supervised learning \cite{aut8, aut28, aut29, aut30, aut33, aut34} has gained significant attention due to its ability to train models without the need for labeled data. Self-supervised learning methods generate pseudo-labels based on the internal structure of the data to guide model learning. It can fully explore the information contained in large amounts of data. Therefore, self-supervised learning is very suitable for pre-training large-scale visual models.

As a branch of self-supervised learning, contrastive learning can also utilize large amounts of unlabeled data for learning. Currently, contrastive learning widely takes advantage of instance discrimination as a pretext task. The core idea is to treat two augmented views of an image as a pair of positive samples, while the augmented views of different images are treated as negative samples. Contrastive learning brings the positive samples close to each other in the embedding space and away from the negative samples. Contrastive learning largely depends on strong data augmentation and a large number of negative samples. However, recent research has also shown that models can learn good representations without negative samples. Nonetheless, existing contrastive learning methods treat each image as a single instance and do not utilize local feature information. This design overlooks the correlation between global and local features in images, which limits the improvement of model generalization to some extent.
\subsection{Additional Memory Structure}
Several studies have explored the use of additional memory structures during model training. In \cite{aut3}, a memory bank is used to store embeddings of all samples. During training, negative embeddings are randomly sampled from it for contrast, loss calculation and encoder parameter update. Finally, the latest encoder updates these sampled embeddings. MoCo maintains a queue to solve the problem that the number of negative embeddings is limited by batch size in another way. This queue stores the latest negative embeddings from several mini-batches. In both of the above methods, the function of additional memory structure is to store a large number of negative embeddings that were originally limited by batch size to improve model contrastive learning performance.

NNCLR \cite{aut9} also maintains an additional queue as a support set. Unlike MoCo, NNCLR uses it to sample nearest neighbors from the dataset in latent space as positive embeddings. MSF \cite{aut32} maintains a queue storing recent embeddings. This queue is used to find the top few nearest neighbors for a given embedding. \cite{aut10} uses the same memory bank as \cite{aut3} to find several neighbors of a given feature as positive embeddings.

However, these methods primarily focus on obtaining more or harder positive and negative embeddings through additional memory structures. They do not consider the diverse effects and corresponding relationships between global and local information in the context of self-supervised learning.
\subsection{Operation on Patches}
There are already many studies \cite{aut12, aut13, aut14} have explored the use of patches in self-supervised learning.

One method is to employ jigsaw puzzles \cite{aut11} as pretext tasks and divide part of an image into patches. After these patches are shuffled, the model predicts their original arrangement. In this process, the network can learn image structure or texture features. Jigsaw clustering \cite{aut15} randomly combines patches from multiple images into new images. Then, the encoder is trained by clustering given patches and restoring original images. Similar to the above researches, \cite{aut16} trains models to learn image features by filling missing regions in randomly masked images.

On the other hand, many self-supervised learning works using transformer \cite{aut37} as encoders also operate on patches. Drawing on experience from the field of natural language processing, ViT \cite{aut38} divides images into patches to match the input requirements of transformer. In MAE \cite{aut18}, certain patches in an image are masked, and the model is trained to reconstruct the pixel information in these masked patches. Similarly, BEIT \cite{aut17} randomly masks some image patches and predicts their corresponding visual tokens. By completing this masked image modeling task, the model can learn the features in the image.

Some of the above researches have also realized that local features play a crucial role in model training. However, most of these methods operate on patches for designing pretext tasks. In our work, we divide images into patches not for pretext tasks but to introduce local feature information and expand the number of samples used for contrast in the network.

\begin{figure*}[t]

\begin{center}
   \includegraphics[width=\linewidth]{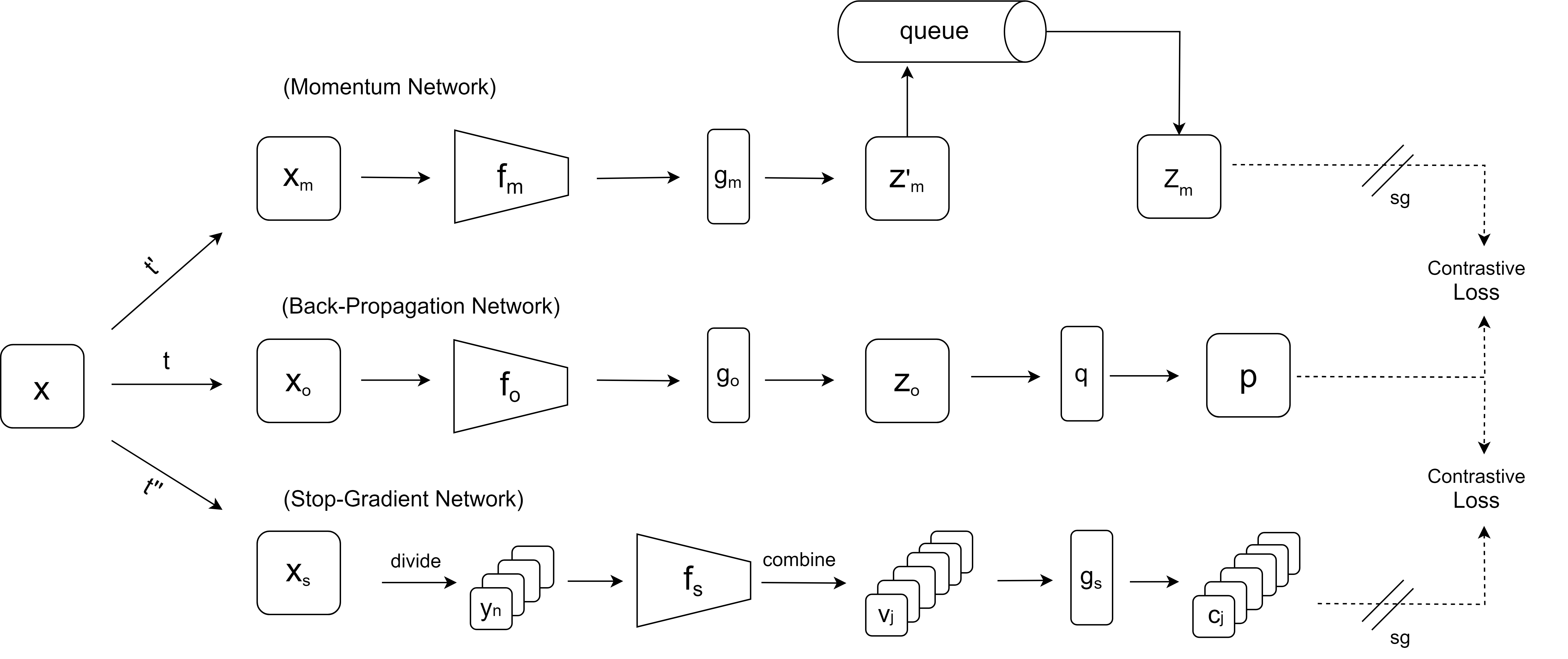}
\end{center} 
   \caption{Overview of MNCLGLF. MNCLGLF includes three networks: only one of them updates parameters by back propagation, and another two update parameters by other policies. With the multi-network framework, MNCLGLF introduces global and local feature information for self-supervised contrastive learning.}
   \label{figure1}
\label{fig:long}
\label{fig:onecol}
\end{figure*}

\section{Approach}
Our goal is to introduce global and local features into self-supervised learning. We propose a multi-network self-supervised learning framework based on global and local features. In this section, we first review previous self-supervised learning in the context of instance discrimination and then we introduce our method in detail. MNCLGLF introduces global and local features for contrastive learning, and comprehensively exploits their visual information correspondence through a multi-network framework. At the same time, it expands the number of samples used for contrast and improves training efficiency.
\subsection{Contrastive Instance Discrimination}
InfoNCE loss \cite{aut3, aut19, aut31} has been one of the most important self-supervised contrastive loss functions since its introduction. The InfoNCE loss is defined as follows:
\begin{equation}
\label{equ1}
    \mathcal{L}_{i}^{InfoNCE} = -log\frac{exp(z_{i}\cdot z_{i}^{+}/\tau)}{exp(z_{i}\cdot z_{i}^{+}/\tau)+ {\textstyle \sum^{B}_{b=1}}exp(z_{i}\cdot z_{b}^{-}/\tau)}
\end{equation}
where $\tau$ is a temperature hyper-parameter, $z_i$ is a given embedding, $z_i^+$ is the corresponding positive embedding, and $z_b^-$ $(b=1, 2, ... , B)$ are $B$ negative embeddings. Therefore, ($z_i$, $z_i^+$) forms a positive pair while ($z_i$, $z_b^-$) forms any negative pair. InfoNCE loss pulls the positive pairs closer together while pushing the negative pairs farther apart in the embedding space.

MoCo has two main contributions. The first is the use of a queue to store a dynamic dictionary, which allows to set a larger dictionary size under limited memory. The second contribution is to slowly update the encoder on the target branch, which ensures that the features in the queue are obtained from the same or similar encoder. These two points provide a large and consistent dynamic dictionary for self-supervised contrastive learning. MoCo V2 \cite{aut4} combines SimCLR’s projector structure and strong data augmentation strategy on the basis of MoCo. Compared to the previous version, MoCo V3 \cite{aut20} removes the queue structure and adds a predictor structure on the online branch by referring to BYOL. The asymmetric structure of the network and the momentum update of the parameters are the key to preventing the model from collapsing in MoCo V3.

\subsection{Our Approach}

We propose a multi-network contrastive learning framework MNCLGLF, as shown in Figure \hyperref[figure1]{1}. The detailed introductions of MNCLGLF are as follows:

\noindent\textbf{Back-Propagation Network.} The back-propagation network consists of a backbone encoder $f_o$, a projector $g_o$, and a predictor $q$. The first augmented view $x_o$ is obtained by performing one random data augmentation on a given image $x$, i.e. $x_o = t(x)$. Based on this view, the backbone encoder and projector output $z_o = g_o(f_o(x_o))$. Then we pass $z_o$ through the predictor $q$ to produce embeddings $p = q(z_o)$. The parameters of this network branch are updated by backpropagation.

\noindent\textbf{Momentum Network.} The momentum network only contains a backbone encoder $f_m$ and a projector $g_m$. The augmented view $x_m$ is obtained by performing the same series of random data augmentation on the given image $x$, i.e. $x_m = t'(x)$. Based on this view, the backbone encoder and projector output a representation $z'_m$ = $g_m(f_m(x_m))$. The parameters of this network branch are updated using momentum as follows: 
\begin{equation}
\theta_m^f = m\theta_m^f + (1 - m)\theta_o^f
\end{equation}
\begin{equation}
\theta_m^g = m\theta_m^g + (1 - m)\theta_o^g
\end{equation}
where $\theta_m^f$ and $\theta_m^g$ are parameters for encoder and projector in the momentum network, $\theta_o^f$ and $\theta_o^g$ are parameters for encoder and projector in the back-propagation network, and $m$ is the momentum coefficient.

In addition, in order to go beyond single instance positives and obtain richer representations, we use the nearest neighbor $z_m$ of the representation $z'_m$ as a positive embedding. We maintain an additional memory structure. Specifically, it is a queue used to retain an embedding set that approximates the distribution of the dataset. The queue is randomly initialized and it will be updated during training phase. At the end of each training step, the oldest batch of embeddings in the queue is dequeued and the current batch of embeddings is enqueued. The size of the queue needs to be set to a sufficiently large value to allow the embedding in the queue to be approximate to the true data distribution.

\noindent\textbf{Stop-Gradient Network.} The stop-gradient network also only contains a backbone encoder $f_s$ and a projector $g_s$. The augmented view $x_s$ is obtained by performing the same series of random data augmentation on the given image $x$, i.e. $x_s$ = $t''$(x). In order to introduce local feature information for self-supervised contrastive learning, inspired from \cite{aut21}, the augmented view $x_s$ is divided into a $2 \times 2$ grid of patches $\left\{y_n|n\in\left\{1, 2, 3, 4\right\}\right\}$. Then the patches are sent to the encoder. However, each patch contains less valid information, which is too difficult for the model to learn. Therefore, we do not directly send each patch’s encoded embedding to subsequent steps. We combine them to obtain $s$ embeddings and $s$ is computed according to the following formula:
\begin{equation}
s = C_{4}^{k}
\end{equation}
where k $\in\left\{1, 2, 3, 4\right\}$ is the number of patches participating in the combination, and $C_{4}^{k} = \frac{4!}{k!(4-k)!}$. Each combined embedding $v_j$ is obtained by averaging the selected $k$ coded embeddings. As a result, we obtain a combined embedding set \textbf{v} = $\left\{v_j|j\in\left\{1, . . . , s\right\}\right\}$. Finally, the combined embedding set \textbf{v} is sent to the projector $g_s$. This network branch finally outputs a group of representations $\left\{c_j|j\in\left\{1, . . . , s\right\}\right\}$. The parameters of the backbone encoder and projector of this network branch are the same as those in the back-propagation network:
\begin{equation}
\theta_s^f = \theta_o^f
\end{equation}
\begin{equation}
\theta_s^g = \theta_o^g
\end{equation}
where $\theta_s^f$ and $\theta_s^g$ are parameters for encoder and projector in the stop-gradient network, $\theta_o^f$ and $\theta_o^g$ are parameters for encoder and projector in the back-propagation network.

By performing the above operations on patches, each combined embedding only contains partial information from the original augmented view. This network branch introduces local feature information into self-supervised contrastive learning and ensures that there is an appropriate information difference between samples and targets. At the same time, the multi-network framework can also enable model to learn correspondence between global feature information and local feature information.

In addition, during division and combination process, this network branch provides several times more positive samples than previous contrastive learning methods. More positive samples can not only improve performance of contrastive representation learning but also improve model training efficiency.

In all three networks above, all data augmentation operations come from same series i.e., $t\sim T$, t$'\sim T$ and t$''\sim T$. Predictors and projectors are both implemented using multi-layer perceptron (MLP). Only the back-propagation network uses back propagation to update parameters while the stop-gradient network parameters stop back propagation update. This is because asymmetric network structure can effectively prevent model training from collapsing, which has been verified as correct in SimSiam. The operations on samples in stop-gradient network and momentum network both play role of semantic perturbation. Finally obtained sample embeddings all come from actual dataset and are very likely new data points not covered by existing data augmentation strategies.

\begin{algorithm*}
\caption{Pytorch-style Pseudocode for MNCLGLF}
\label{alg1}
\SetAlgoLined
    
    \PyComment{f$\_$o: the back-propagation network [encoder, projector, predictor]} \\
    \PyComment{f$\_$s: the stop-gradient network [encoder, projector]} \\
    \PyComment{f$\_$m: the momentum network [encoder, projector]} \\
    \PyComment{m: momentum coefficient, Q: queue} \\
    \PyComment{$\lambda$: balancing hyperparameter, $\tau$: temperature} \\
    \PyCode{ } \\
    \PyCode{for x in loader:} \PyComment{load a minibatch} \\
    \Indp   
        \PyCode{x1, x2, x3 = aug(x), aug(x), aug(x)}\\
        \PyComment{random augemtation, NxCxHxW}\\
        \PyCode{ } \\
        \PyCode{p1, p2, p3 = f$\_$o(x1), f$\_$o(x2), f$\_$o(x3)}\\
        \PyComment{the back-propagation network encode, project, predict}\\
        \PyCode{z1, z2 = f$\_$m(x1), f$\_$m(x2)}\\
        \PyComment{the momentum network encode, project}\\
        \PyCode{z1, z2 = NN(Q, z1), NN(Q, z2)}\\
        \PyComment{top-1 NN}\\
        \PyCode{ } \\
        \PyCode{y$\_$1, y$\_$3 = divide(x1), divide(x3)}\\
        \PyComment{divide step, 4NxCx(H/2)x(W/2)}\\
        \PyCode{vd$\_$1, vd$\_$3 = f$\_$s[0](y1), f$\_$s[0](y3)} \\
        \PyComment{the stop-gradient network encode}\\
        \PyCode{v$\_$1, v$\_$3 = combine(vd$\_$1), combine(vd$\_$3)} 
        \PyComment{combine step}\\
        \PyCode{c$\_$1, c$\_$3 = f$\_$s[1](v$\_$1), f$\_$s[1](v$\_$3)} 
        \PyComment{project}\\
        \PyCode{ } \\
        \PyCode{loss$\_$s = (L(c$\_$1, p3) + L(c$\_$3, p1)) / 2} \\
        \PyCode{loss$\_$m = (L(z1, p2) + L(z2, p1)) / 2} \\
        \PyCode{loss$\_$total = loss$\_$s + $\lambda$ * loss$\_$m} \\
        \PyCode{loss$\_$total.backward()} 
        \PyComment{weight update}\\
        \PyCode{update(f$\_$q.params)} \\
        \PyCode{f$\_$m.params = m * f$\_$m.params + (1 - m) * f$\_$q[:2].params} \\
        \PyCode{f$\_$s.params = f$\_$q[:2].params} \\
        \PyCode{update$\_$queue(Q, z1)} \\
        \PyCode{ } \\
        \PyCode{ } \\
    \Indm 
    \PyCode{def NN(Q, z):}\\
    \Indp
        \PyCode{Q, z = normalize(Q, dim=1), normalize(z, dim=1)} 
        \PyComment{l2-normalize}\\
        \PyCode{sim = mm(z, Q.t())} \\
        \PyCode{index = sim.argmax(dim=1)} \\
        \PyComment{top-1 NN indices}\\
        \PyCode{return Q[index]} \\
        \PyCode{ } \\
        \PyCode{ } \\
    \Indm
    \PyCode{def L(h, p):}
    \PyComment{contrastive loss} \\
        \Indp
        \PyCode{h, p = normalize(h, dim=1), normalize(p, dim=1)}
        \PyComment{l2-normalize}\\
        \PyCode{h = h.split(p.shape[0])} \\
        \PyComment{calculate loss}\\
        \PyCode{loss = 0} \\
        \PyCode{for $\_$h in h:} \\
        \Indp
            \PyCode{logits = mm($\_$h, p.t())}\\
            \PyCode{labels = range(p.shape[0])}\\
            \PyCode{loss += CorssEntropyLoss(logits/$\tau$, labels)}\\
        \Indm
        \PyCode{return loss /= len(h)}\\
    \Indm
    \PyCode{ } \\
\end{algorithm*}

\noindent\textbf{Contrastive loss function.} The contrastive loss function used in MNCLGLF is a variant of InfoNCE (\hyperref[equ1]{1}) with only minor differences compared to it. The embeddings passed into contrastive loss function may be multiple combined embeddings provided by stop-gradient network branch. Therefore, we average multiple contrastive loss values formed by these combined embeddings with embeddings provided by the back-propagation network branch. The overall contrastive loss function of MNCLGLF is as follows:
\begin{equation}
\label{equ2}
    loss_{total} = loss_s + \lambda loss_m
\end{equation}

where $loss_s$ is the contrastive loss between the stop-gradient network and the back-propagation network, and $loss_m$ is the contrastive loss between the momentum network and the back-propagation network. $\lambda$ $> 0$ is a balancing hyperparameter.

The overall procedure of MNCLGLF is summarized in Algorithm \hyperref[alg1]{1}.

\section{Experiments}
\subsection{Implementation Details}
\noindent\textbf{Datasets and Device Performance.} Typically, most self-supervised learning methods are validated on ImageNet \cite{aut39} using larger batch sizes. This is too demanding for experimental equipment. Limited by the experimental environment, MNCLGLF is compared with other methods on three benchmark datasets, including CIFAR10 \cite{aut22}, CIFAR100 \cite{aut22} and TinyImageNet \cite{aut23}. Table \hyperref[tab1]{1} shows the details of these datasets.

For CIFAR10 and CIFAR100, we use training set for pre-training and fine-tuning downstream tasks, and employ test set for evaluation. As for TinyImageNet, we use training set for pre-training and fine-tuning downstream tasks, and utilize 10K validation set instead of the original test set for evaluation.

All experiments in this paper are trained and evaluated on Nvidia GTX 3090 GPU. All methods employ the following configurations.

\begin{table}[H]
\caption{The details of datasets.}
\label{tab1}
\begin{center}
\scalebox{0.8}{
\begin{tabular}{ccccc}
\hline
Dataset      & Number of classes & Image size & Training set & Test set \\ \hline
CIFAR10      & 10                & 32 × 32    & 50000        & 10000    \\
CIFAR100     & 100               & 32 × 32    & 50000        & 10000    \\
TinyImageNet & 200               & 64 × 64    & 100000       & 10000    \\ \hline
\end{tabular}
}
\end{center}
\end{table}

\noindent\textbf{Architecture.} We use ResNet-18 \cite{aut24} as the backbone encoder and MLP to implement the projector and predictor. The architecture of the projector is three fully connected layers with sizes of [512, 2048], [2048, 2048] and [2048, 2048] respectively. After each fully connected layer, a batch-normalization (BN) \cite{aut25} layer follows. Additionally, after the first and second BN layers, a rectified linear unit (ReLU) layer is applied.

The architecture of the predictor is two fully connected layers with sizes of [2048, 512] and [512, 2048] respectively. BN and ReLU are located between the two fully connected layers.

\noindent\textbf{Pre-Training.} Table \hyperref[tab2]{2} shows the detailed configurations of self-supervised pre-training. Regarding data augmentation, we employ various techniques to increase the diversity of training data. Random cropping is applied to introduce spatial variability, while random horizontal flipping further augments the dataset. Furthermore, we employ color distortion and Gaussian blur to facilitate better feature learning.

\begin{table}[H]
\caption{The details of configurations.}
\label{tab2}
\begin{center}
\scalebox{1.0}{
\begin{tabular}{c|c}
\hline
Configurations                          & Values \\ \hline
Epochs                                  & 200    \\
Optimizer                               & SGD \cite{aut26}    \\
Optimizer momentum                      & 0.9    \\
Weight decay                            & 1e-4   \\
Learning rate schedule                  & Cosine \\
Initial Learning rate                   & 0.1    \\
Final Learning rate                     & 0      \\
Batch size                              & 256    \\
Temperature $\tau$                           & 1.0    \\
Momentum coefficient $m$   & 0.99   \\
The size of additional memory structure & 16384  \\ \hline
\end{tabular}
}
\end{center}
\end{table}

\noindent\textbf{Evaluation.} We use the same evaluation strategy as SimCLR, BYOL and SimSiam. We freeze the parameters of the pre-trained backbone encoder $f_o$. Then add a linear classifier of size [512, $class_{num}$], where $class_{num}$ is the number of classes in the dataset. The performance of the model is evaluated by the accuracy of linear classification. Linear evaluation uses layer-wise adaptive rate scaling (LARS) optimizer \cite{aut27} with momentum of 0.9 and weight decay of 0. We utilize a cosine decay schedule from 0.8 to 0 to adjust the learning rate and train for 90 epochs. Batch size is 256.

\subsection{Linear Evaluation}
The comparison of linear evaluation results with other methods is shown in Table \hyperref[tab3]{3}. Limited by the device, the batch size is set to 256, the size of all additional queues is 16384, and the temperature is set to 1. We use top-1 accuracy and top-5 accuracy to evaluate the performance of all methods on these three benchmark datasets. Top-1 accuracy refers to the accuracy rate that the class with the highest probability is the correct class. Top-5 accuracy refers to the accuracy rate that the top 5 classes with the highest probability contain the correct class. MNCLGLF outperforms other classical self-supervised learning methods on three benchmark datasets. The combination of global and local feature information can help our model learn better representations and enhance the generalization ability of the model. However, compared with supervised methods, MNCLGLF still has a gap in accuracy.

\begin{table}[H]
\caption{Linear evaluation results. The optimal results are shown in bold. Compared to some existing classical self-supervised learning methods, MNCLGLF achieves the best results on all three benchmark datasets. The results of linear evaluation prove that MNCLGLF is effective.}
\label{tab3}
\begin{center}
\scalebox{0.9}{
\centering
\begin{tabular}{ccccccc}
\hline
Methods & \multicolumn{2}{c}{CIFAR10}     & \multicolumn{2}{c}{CIFAR100}   & \multicolumn{2}{c}{TinyImageNet} \\
                         & top1           & top5           & top1          & top5           & top1           & top5            \\ \cline{2-7} 
Supervised               &92.89      &99.84       &71.93            &92.21             &54.03          &77.02           \\
MoCo V2 \cite{aut4}                  & 84.03          & 99.53          & 50.49         & 80.28          & 26.01          & 51.92           \\
SimSiam \cite{aut6}                 & 84.16          & 99.45          & 48.19         & 78.41          & 25.28          & 50.63           \\
SimCLR \cite{aut1}                  & 87.11          & 99.64          & 59.20         & 86.41          & 31.39          & 59.08            \\
BYOL \cite{aut5}                    & 85.58          & 99.58          & 57.12         & 84.83          & 32.76          & 59.55           \\
NNCLR \cite{aut9}                    & 87.66          & 99.63          & 60.91         & 87.50          & 40.46          & 67.05           \\
Fast-MoCo \cite{aut21}               & 85.93          & 99.57          & 61.39         & 87.31          & 39.91          & 66.26           \\
\textbf{MNCLGLF}            & \textbf{90.18} & \textbf{99.78} & \textbf{65.20} & \textbf{89.72} & \textbf{41.70}  & \textbf{68.36}  \\ \hline
\end{tabular}
}
\end{center}

\end{table}

\subsection{Ablations}

\noindent\textbf{Balancing hyperparameter $\lambda$.} We explore the effect of balancing hyperparameter $\lambda$ in Equation \hyperref[equ2]{7} as shown in Table \hyperref[tab4]{4}. This experiment was conducted on CIFAR10 dataset with other hyperparameters unchanged except for $\lambda$. The table shows the top-1 accuracy. When $\lambda$ equals 6, our model achieves best performance. Since the downstream task is image classification, the global feature information has a large impact on model performance. So when $\lambda$ is a smaller value, the model performance is not satisfactory. However, when $\lambda$ is too large, the contrast between the stop-gradient network and the back-propagation network accounts for a small proportion. The model will suffer from performance reduction due to excessive neglect of local feature information. Therefore, only by setting $\lambda$ to a moderate value, our model can make full use of global and local feature information for learning.

\begin{table}[H]
\caption{Effect of different $\lambda$ for MNCLGLF.}
\label{tab4}
\begin{center}
\scalebox{1}{
\centering
\begin{tabular}{cccccc}
\hline
$\lambda$    & 2     & 4    & 6     & 8     & 10    \\ \hline
MNCLGLF & 89.33 & 89.50 & \textbf{90.18} & 89.26 & 89.27 \\ \hline
\end{tabular}
}
\end{center}
\end{table}

\noindent\textbf{Number of samples used for contrast.} We explore the effect of more samples on training efficiency as shown in Table \hyperref[tab5]{5}. We only randomly select one from multiple combined embeddings obtained from stop-gradient network branch as a positive embedding for contrast. Then we compare linear evaluation results every 50 epochs on CIFAR10 with NNCLR and MNCLGLF. The table shows the top-1 accuracy.

The multi-network framework provides more samples for contrast. Therefore, even with only one combined embedding, MNCLGLF trains faster than NNCLR. The training efficiency of MNCLGLF is further improved when we use more positive pairs provided by the divide operation on the base of the multi-network framework. When the model is trained to the 150th epoch, the accuracy of MNCLGLF reaches 90.17\%, while the performance of the method with only one combined embedding still falls short of its optimal performance. At the same time, the linear evaluation of NNCLR is significantly different from its optimal performance 87.66\%. In addition, the linear evaluation results of MNCLGLF at the 100th epoch of model training are already comparable to the performance of NNCLR at the 200th epoch of model training.

This result proves that our method’s operation of increasing number of positive pairs used for contrast during training process can improve performance and efficiency of self-supervised contrastive learning.

\begin{table}[H]
\caption{Effect of different number of samples for training efficiency.}
\label{tab5}
\begin{center}
\scalebox{0.9}{
\begin{tabular}{ccccc}
\hline
Methods                  & 50e   & 100e  & 150e  & 200e  \\ \hline
NNCLR \cite{aut9}                    & 72.48 & 82.54 & 86.86 & 87.66 \\
MNCLGLF(one embedding) & 80.75  & 87.47 & 89.36 & 89.84 \\
MNCLGLF                     & 80.55 & 87.77 & 90.17 & \textbf{90.18} \\ \hline
\end{tabular}
}
\end{center}
\end{table}

\noindent\textbf{Number of encoded embeddings participating in the combination.} We explore the effect of the number of encoded embeddings participating in the combination as shown in Table \hyperref[tab6]{6}. Where $k$ is the number of encoded embeddings participating in the combination. In the stop-gradient network branch, we vary the number of encoded embeddings that participate in the combining operation. Then we perform linear evaluation on the CIFAR10 dataset. As we expected, when the encoded embeddings of the patches were sent directly to the projector without being combined, the model did not perform perfectly. This is because each patch contains less effective information, which is too difficult for the model to learn. And when there are more encoding embeddings participating in the combination, the linear evaluation performance will also decrease. This is due to the fact that the embeddings output by the branches of the network have less information difference between them. Therefore, the model can only show good performance when each embedding contains a moderate amount of original image information.

\begin{table}[H]
\caption{Effect of different number of encoded embeddings participating in the combination.}
\label{tab6}
\begin{center}
\scalebox{1.0}{
\begin{tabular}{ccc}
\hline
$k$ & top-1 & top-5 \\ \hline
1 & 89.54 & 99.73 \\
2 & \textbf{90.18} & \textbf{99.78} \\
3 & 89.79 & 99.73 \\
4 & 89.61 & 99.72 \\ \hline
\end{tabular}
}
\end{center}
\end{table}

\section{Conclusion}
We introduce global and local feature information for self-supervised contrastive learning through multiple networks and improve model performance in downstream tasks. On three benchmark datasets, MNCLGLF outperforms some existing classical self-supervised contrastive learning methods. At the same time, while keeping training data unchanged, we expand the number of samples used for contrast, generating more positive pairs and improving the training efficiency of the model. However, these additional samples are not equally important for the learning of the model. In future work, we hope to find a way to effectively distinguish more important samples.

\section*{Acknowledgement}
This work was supported by the National Natural Science Foundation of China under Grant No. 61906098.


\bibliographystyle{ieeetr}
\bibliography{example}

\end{document}